\documentclass[letterpaper, 10 pt, conference]{ieeeconf}  % Comment this line out if you need a4paper

\IEEEoverridecommandlockouts                              % This command is only needed if 
\usepackage{graphicx}
\usepackage{booktabs}
\usepackage{tabularx}
\usepackage{makecell}
\usepackage[table]{xcolor} 
\usepackage{float}
\usepackage{caption}
\usepackage[hidelinks]{hyperref}
 \usepackage{amsmath}

\definecolor{gold}{RGB}{255,215,0} 
\definecolor{silver}{RGB}{192,192,192} 
\definecolor{bronze}{RGB}{205,127,50}
\definecolor{blue}{RGB}{66, 173, 245}
\title{\LARGE \bf
CSIR: Contextually and Socially Informed Robot for Efficient Person Goal Navigation
}

\author{Tyler Chung$^{1}$, Nahl Farhan$^{2}$, Hao Zhang$^{3}$, Mingfeng Yuan$^{4}$, Jinjun Shan$^{3}$, Amy Wu$^{5}$, Yue Hu$^{1}$% <-this % stops a space
\thanks{$^{1}$T. Chung and Y. Hu are with the Department
of Mechanical and Mechatronics Engineering, University of Waterloo, Waterloo, ON, Canada
        {\tt\small \{tyler.chung, yue.hu\}@uwaterloo.ca}}%
\thanks{$^{2}$N. Farhan is with the Department of Mechatronics Engineering, Ontario Tech University, Oshawa, ON, Canada
        {\tt\small \{nahl.farhan\}@ontariotechu.net}}%
\thanks{$^{3}$H. Zhang and J. Shan a the Department of Earth and Space Science Engineering, York University, Toronto, ON, Canada
        {\tt\small \{zhang85, jjshan\}@yorku.ca}}%
\thanks{$^{4}$M. Yuan is the University of Toronto Institute for Aerospace Studies and the University of Toronto Robotics Institute, University of Toronto, Toronto, ON, Canada
        {\tt\small \{mingfeng.yuan\}@robotics.utias.utoronto.ca}}%
\thanks{$^{5}$A. Wu is at the Department of Mechanical and Materials Engineering, Queen's University, Kingston, ON, Canada
        {\tt\small \{amy.wu\}@queensu.ca}}%
}

\begin{document}

\maketitle
\thispagestyle{empty}
\pagestyle{empty}

%%%%%%%%%%%%%%%%%%%%%%%%%%%%%%%%%%%%%%%%%%%%%%%%%%%%%%%%%%%%%%%%%%%%%%%%%%%%%%%%
\begin{abstract}

We address the Person Goal Navigation (PersonNav) problem, enabling robots to search for people under realistic constraints on information accessibility and language uncertainty. This tackles the current solutions revolving around rigid person finding systems requiring exact knowledge of the individual for item-delivery in indoor settings. While the focus in literature is on learned methods using unavailable public or sparse data due to human privacy. We propose a planning framework that combines distance and semantic information about the recipient (habits and intent) weighted by the trust of the user-provided information. A synthetic benchmark of scenarios is also developed including actors, items, and requests to evaluate performance before real-world deployment aiming to simulate natural language human-robot interactions. 
% We propose a planning framework that combines distance and semantic scores, weighted by the trust of user-provided information, and introduce a synthetic benchmark of actors, items, and requests to evaluate search performance. 
Results show our informed search outperforms classical distance-based graph baselines, while semantics alone lead to ungrounded, sporadic search. Our method achieves strong gains over baselines, with an LLM-based variant performing comparably, and its explicit belief representation naturally supports future Bayesian filtering. Hardware tests demonstrate our method supports real-world embodiment able to leverage between semantic and distance information in a real setting. This work moves toward more intelligent mobile service agents capable of human-like, informed search in realistic environments to be leveraged in day-to-day use. 
\textcolor{red}{\href{https://anonymous.4open.science/r/personnav-site-4ED2/index.html}{https://anonymous.4open.science/r/personnav-site-4ED2/index.html}.} 

%\href{https://chungtyler.github.io/personnav-site/}{See Demos Here}.

\end{abstract}

%%%%%%%%%%%%%%%%%%%%%%%%%%%%%%%%%%%%%%%%%%%%%%%%%%%%%%%%%%%%%%%%%%%%%%%%%%%%%%%%

\section{INTRODUCTION}
About one in three physicians experience burnout interfering with not only their well being but ability to provide quality care for others \cite{De_Hert_2020}. As result of the working conditions, this has led to an increase in burnout, health issues, and errors in medical practice \cite{CIHI}. To alleviate staffing pressure, nursing robots are used to automate item-delivery tasks \cite{wang2024advancing}, with platforms such as MOXI or TUG use voice or graphical interfaces to schedule and design routes for transportation in hospitals \cite{wang2024advancing}. One limitation is that current tools require fixed location scheduling to deliver items, but people are dynamic, not bounded to locations as they are goal driven. This realizes the problem that these solutions lack the capability to perform its tasked when the assumption that people are static fail. Moving away from fixed location settings and towards dynamic goal-driven people raises the question of “How can robots find people given partial observations of their behaviours and the environment”, we define this problem space as Person Goal Navigation (PersonNav). While this is motivated by the healthcare landscape, this problem expands to other tasks ranging from food delivery, office supply transportation, to nursing home monitoring.

This paper proposes the development of a travel efficient PersonNav framework for indoor item-delivery tasks under limited and uncertain information about the recipient (whom we want to deliver to). Our proposed task space for PersonNav is catered to real-world feasibility in indoor item-delivery scenarios, where a user requests the agent to deliver an item to a recipient (consented information to the robot). Current methods in literature require critical pieces of information from human-activity logs \cite{tipaldi2011want} or daily schedules \cite{fung2025mllm}, but are partially or fully unavailable in the real-world due to privacy reasons. The current gap is that existing methods leverage critical and sensitive information without consideration of realistic information availability.  To address this problem, our work aimed to investigate methods to extrapolate a person’s location from natural language requests and handle information uncertainty (e.g. Low: “They might be at ...” vs High: “They’re definitely at ...”). We hypothesized that by using several weaker/indirect signals of the person’s whereabouts, we can improve search under uncertain information gains for efficient indoor item-delivery tasks while abiding by real-world constraints in human privacy. The contributions of this paper are:
% \begin{itemize}
%     \item  Contextually and Socially Informed Robots (CSIR): an efficient PersonNav framework under information uncertainty for item-delivery tasks;
%     \item A synthetic benchmark with meaningful scenarios, actors, and ground truths; and
%     \item An evaluation against relevant graph-based and semantic baselines.
% \end{itemize}

\begin{itemize}
    \item \textbf{Contextually and Socially Informed Robots (CSIR)}, that addresses the PersonNav problem aimed to perform efficiency item-delivery tasks under varying linguistic uncertainty.
    \item A synthetic benchmark to generate meaningful scenarios, actors, and ground truths for human-robot interactions
    \item An evaluation on the benchmark and methods using relevant graph-based and semantic baselines across 3 different simulated environments (1,500 scenarios) alongside real-world hardware tests on a mobile robot
\end{itemize}

%The objectives the work aimed to cover are as follows:

% \begin{itemize}
%     \item[O1)] Develop efficient PersonNav frameworks under information uncertainty for item delivery tasks
%     \item[O2)] Generate meaningful scenarios and actors with ground truths for benchmarking
%     \item[O3)] Evaluate framework performance against relevant graph-based and semantic baselines
% \end{itemize}

% \begin{figure}[t]
%     \centering
%     \includegraphics[width=0.9\linewidth]{figures/problem_formulation_clip.png}
%     \caption{Problem Finding Task Office Example}
%     \label{fig:teaser}
% \end{figure}

\section{RELATED WORKS}
\subsection{Object Goal Navigation (ObjectNav)}
The PersonNav problem space is inspired by the Object Goal Navigation (ObjectNav) field where an agent navigates in an unseen environment to find a requested object \cite{sun2024survey}. The methods deployed can range from end-to-end methods for navigating the environment to zero-shot methods that leverage pre-trained semantic models to inform search such as Large Language Models (LLMs) \cite{sun2024survey}. LLMs are trained on a massive corpus of human text information and hence developed a causal understanding of the human world semantically \cite{ma2025causal}. This helps informing search by leveraging priors using zero-shot methods to infer object locations (e.g. food can normally be found in the kitchen). The difference between ObjectNav and PersonNav is that the former focuses on static object search while the latter on goal-driven people search requiring a different set of observations to inform search.

\subsection{Probabilistic and Semantic-Based Person Finding}
Literature on person finding tasks exploits information such as activity logs or schedules and fall under the categories of behaviour-based search, spatio-temporal search, and semantic-based exploration. Behavioural search uses a human activity schedule (probabilities of being in different locations at a given time). This frames the problem as a partially observable Markov decision process (POMDP) \cite{cassandra1998survey} and uses a learned policy leveraging activity, last known location, distance, and field-of-view to explore optimal frontiers \cite{mehdi2014behavior}. For spatio-temporal search, \cite{tipaldi2011want} uses spatial affordance maps to learn where a person is using a Poisson distribution which maps human activity during the day to optimize encounters to the person in an office environment. Semantic-based exploration is similar to ObjectNav methods, using the person's behaviour and schedule through natural language prompts to explore a set of waypoints on a map to search for a recipient \cite{fung2025mllm}. In general these methods requires information such as human-activity logs or daily schedules that are not fully or at all available in the real-world for privacy concerns, making real-world deployment in public settings challenging. Therefore, further methods are required to extrapolate indirect observations of a person's behaviour to inform person search while under ethical constraints.

\section{PROBLEM FORMULATION}
\label{problem_formulation}
We formulate PersonNav as  an item-delivery scenario where an agent is tasked by a user with finding and delivering an item to a recipient whose current location is unknown. The environment is represented as a graph of nodes called locations, where a recipient resides in one of these locations. The agent's objective is to determine which locations to search and in what order to find the recipient in the shortest distance to maximize travel efficiency. The agent is provided with an annotated map (location labels and descriptions), a list of recipient names and roles/occupations (e.g. employees in an office), and a request (user provided information about the recipient). This is designed to represent minimal but available factual information in workplace settings.

The request takes the form of a natural language prompt (e.g "Hey can you deliver the pen to Jim? He might be at the office desk."). It specifies the item to be delivered, recipient, and hints about the recipient's location with varying uncertainty (degree of belief from the user). The following types of hints are considered (see Table \ref{tab:hint_types} for examples): %in \ref{tab:hint_types} with examples showing the request. 
%This is to cover spatio-temporal information that is delivered in a plausibly realistic manner from the requester.
\begin{itemize}
    \item \textbf{None} - Defines no hinted information other than the item and recipient name/occupation;
    \item \textbf{Recency} - Defines the last seen location of the person with informational uncertainty based on the time elapsed since the observation;
    \item \textbf{Behavioural} - Defines a prediction of where the recipient is based on the user's understanding of the recipient's habits and routine;
    \item \textbf{Contextual} - Defines a prediction of where the recipient is based on the user's understanding of the recipient's reason to need/use the item;
    \item \textbf{Behav. \& Context.} - Defines a prediction of where the recipient is based on the user's understanding of both the recipient's routine and intent of using the item.
\end{itemize}

\begin{table}[h]
    \centering
    \caption{Hint Type Taxonomy and Examples}
    \label{tab:hint_types}
    \begin{tabularx}{\columnwidth}{
        >{\raggedright\arraybackslash\hsize=0.5\hsize}X 
        >{\raggedright\arraybackslash\hsize=0.5\hsize}X 
        >{\raggedright\arraybackslash\hsize=2.0\hsize}X
    }\toprule
         \textbf{Type} & \textbf{Certainty} & \textbf{Example} \\\midrule
         None & - & Hey can you give this screwdriver to Jane? \\
         Recency & High & This hammer is for Deshawn, he was right over by the lounge just a brief moment ago. \\
         Behavioural & Low & Give this folder to Martin, not too sure but they're sometimes at the lab. \\
         Contextual & Low & Can you give Ramirez the whistle. He may need it for running drills today, but can't say for certain, so try the field. \\
         Behav. \& Context. & High & Could you bring this laptop to Sana? She's normally in the studio editing, and she'll definitely need it for that. \\ \bottomrule
    \end{tabularx}
\end{table}

%This problem formulation defines the task space, what information is provided, and what it contains to meaningfully design experiments and ablations while basing it in plausible real-world observations for the agent.

\section{CSIR FRAMEWORK}
\begin{figure*}[t]
    \centering
    \includegraphics[width=1\linewidth]{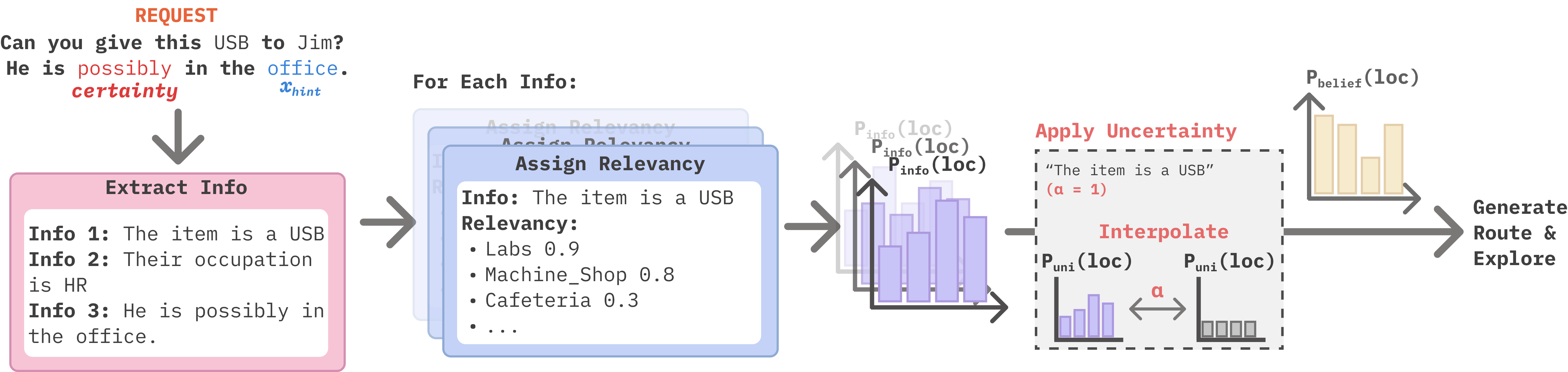}
    \caption{CSIR Framework Belief Generator}
    \label{fig:framework}
\end{figure*}

Our proposed framework, CSIR (Contextually and Socially Informed Robot), leverages both semantic information about the recipient and environment alongside the geometric paths between locations to solve the PersonNav problem in item delivery tasks. This is achieved by using the user request to generate multiple hypotheses of where the recipient may be based on information extracted from the request  (e.g. the need for the item, recipient occupation, and hinted information). Each hypothesis is represented as a probability distribution over the recipient's possible locations. Each distribution is modulated based on their level of uncertainty (derived from the speaker's belief) to represent trust associated with that information, inspired by Bayesian filtering techniques such as the Kalman Filter, which modulates sensor observations based on their uncertainty \cite{5158765}. The distributions are then multiplied to produce a final distribution representing the agent's initial belief. The route is then generated by solving a weighted version of the Travelling Repairman Problem (TRP) \cite{afrati1986complexity} using the initial belief as weights and distances between each location. This generates an optimal route aimed at minimizing the cumulative belief-weighted travel distance at every location along the route. %This then solves for an optimal route aimed at minimizing the cumulative belief-weighted travel distance. 
The agent then visits each location (the coordinates of the entrance door of a room/space) until the recipient is found.

\subsection{Belief Generator}
\label{belief_generator}

The belief generator is used to generate the initial belief as a probability distribution of finding a recipient for all locations given the provided information defined in Section \ref{problem_formulation}. The request is first parsed by an LLM, which extracts individual pieces of information that contribute to estimating where the recipient could be, as shown in Fig. \ref{fig:framework}. The LLM outputs a JSON schema containing a list of the extracted information, its type (PERSON, ITEM, or HINT), and rationale for its extraction to provide transparency in its decision-making. For each extracted piece of information, another LLM is used to assign relevancy scores \(S(loc)\in[0, 1]\) for each location \(loc\) by comparing the information to the descriptions of each location. The LLM calls per information is done in parallel to reduce inference time during planning. %This is done as LLMs are capable of providing semantic meaning relative to each other even if its values are not fully calibrated \cite{} (this provides enough meaning in our initial belief generation). Hence, after generating t
These scores capture the semantic relevance of each location and  are converted into valid probability distributions \(p_{info}(loc)\) using linear normalization, denoting the hypothesis of the recipient's location for each piece of information. %As a note commonly in machine learning other normalization techniques are used such as Softmax \cite{} but distort the relative meaning between labels as it's best for learning one-hot vector representations (does not apply in our case).

\begin{equation}
    p_{info}(loc) = \frac{S(loc)}{\sum^n_{i=1}{S(loc_i)}}
    \label{linear_norm}
\end{equation}

For each hypothesis, a certainty weight \(\alpha \in [0,1]\) represents the trust associated with the information, where \(1.0\) is high and \(0.0\) is low certainty. This allows information expressed in natural language to be weighted according to its certainty rather than treated as absolute ground truth \cite{pmlr-v205-ichter23a}. %This is to allow the agent to decide how much trust should be applied to each observation (as mentioned similar to Bayesian-filtering methods modulating sensor observations on uncertainty/variance). This is so information even as natural language should not be taken always at face value but with varying degrees of uncertainty (addressing how common robotic natural language tasks take prompts as absolute information/ground truth \cite{}). 
The item and the recipient occupation are treated as factual as in Section \ref{problem_formulation}, so $\alpha = 1$, while hints represent the main source of uncertainty. For hints, an uncertainty predictor estimates $\alpha$ from the user's expressed confidence, including hedging language or epistemic modality \cite{vold2006epistemic}. A transformer-based semantic model is used (e.g. encoder-based transformer models such as BERT \cite{gardazi2025bert}) to map this language to a continuous certainty value. %This is done in similar fashion to sentiment analysis for reviews assigning high or low buckets of rating to comments or reviews \cite{} except a more continuous value is required to modulate uncertainty of the hypothesis. 

% Note not sure if this should be explained in this section since it's about how the module is trained 
The uncertainty predictor is trained on a synthetically generated dataset that contains data of different statements of varying uncertainty (e.g. "I can't say for certain but it might rain today." vs "I'm confident it will rain today") modulated by hedging language into 5 categories (HIGH, MODERATE\_HIGH, NEUTRAL, MODERATE\_LOW, LOW). These labels are mapped to values in \([0, 1]\) where \(1.0\) is HIGH and \(0.0\) is LOW, providing a continuous certainty weight $\alpha$.

%As a note hedging language is used as an indicator of uncertainty which assumes the more confident the user the greater certainty of information (this is not always true but it's assumed the user will mention information with higher tones of confidence if their answers are more correct), \cite{} shows how language uncertainty can be numerically bucketed into a range of \([0, 100\) meaningfully therefore representing how language displays meaningful values in uncertainty.

The certainty weight\(\alpha\) is then used to interpolate between the information-derived hypothesis \(p_{info}(loc)\) representing the highest information gain, and the uniform distribution \(p_{uni}(loc)\) as no information gain:%. By linear interpolation between the distributions using the weight this modulates how much informational gain should be leveraged from the information or hint using the following equation:
\begin{equation}
    p(loc) = \alpha p_{info}(loc) + (1 - \alpha) p_{uni}(loc)
    \label{linear_interp}
\end{equation}

Finally, the resulting hypotheses are combined by multiplying each distribution together and normalizing. %\(\frac{p(loc|info_1)p(loc|info_2)p(loc|info_3)}{\eta}\), where \(\eta\) is the normalization factor. 
This reinforces locations supported by agreeing hypotheses while diminishing those with conflicting evidence \cite{hinton2002training}. The resulting distribution forms the agent's initial belief for route planning. %is to allow agreeing hypothesis for a location to stay while disagreeing hypothesis for a given location will diminish. After computing this becomes the generated initial belief of the agent it will use for planning its route.

\subsection{Route Planner}

After generating the initial belief of where the recipient is given the information, the route planner's objective is to balance the relevance of each location with the distance required to reach it, reducing the total travel distance before finding the recipient. Different methods can be used to generate the route (a list of ordered locations to visit) such as heuristic or optimal approaches. We adopt a dynamic programming approach to solve the Travelling Repairman Problem (TRP) \cite{afrati1986complexity}, derived from the Held-Karp dynamic programming solution for the Travelling Salesman Problem (TSP) \cite{held1962dynamic}). 
%As background the problem-space is formalized as a series of nodes or locations with edges defining the cost to reach that location from a given current location. For a problem with \(n\) locations the TSP aims to minimize the global shortest Hamiltonian path (meaning each location is visited at least once). The objective for the TSP is formalized as \(min\sum^{n}_{i=1}\sum^{n}_{j=1} c_{ij} x_{ij}\), where \(n\) is the number of locations, \(c_{ij}\) is the distance between location i and j, and \(x_{ij}\in \{0,1\}\) is a binary value if the route goes between location i to j (this is essentially the route solvers aim to optimize). The dynamic programming approach recursively computes every possible combination of path using memory caching to reduce repeated computations to improve compute. Since the problem is NP-hard (non-deterministic polynomial time) the exact solution may not feasibly be computed due to time and resource depending on the number of nodes (this is later evaluated in the results \ref{}). 

Unlike TSP, which minimizes the total distance required to visit all locations, TRP aims to minimize the global cumulative cost (in this case distance) for each node it visits, this can also be weighted to give priority to nodes called Weighted TRP or WTRP. The objective for WTRP is as follows: \(min\sum^{n}_{i=1} w_iC_i\), where \(n\) is the number of locations, \(w_i\) is the probability of finding the recipient at i, and \(C_i\) is the cumulative distance to reach i. The initial belief generated in Section \ref{belief_generator} therefore provides the location weights, prioritizing likely recipient locations while accounting for travel distance.

% The WTRP solver also uses a dynamic programming approach to get the exact solution similar to the Held-Karp solution traversing every combination of locations to compute the cumulative cost. 
Since the WTRP uses a dynamic programming approach for the exact solution its time complexity is \(O(n^22^n)\), which is tractable for the number of locations explored in this paper (e.g. \(<20 \text{ locations}\)), as evaluated in Section \ref{sim_result}. %The TRP in this case make it suitable for the problem space as it allows for weighted planning based on the belief aiming to minimize the cumulative cost between each location it traverse to or rephrased as aiming to maximize its encounter with the recipient based on weighted priority.

\label{scenario_generator}
\begin{figure*}[t]
    \centering
    \includegraphics[width=1.0\linewidth]{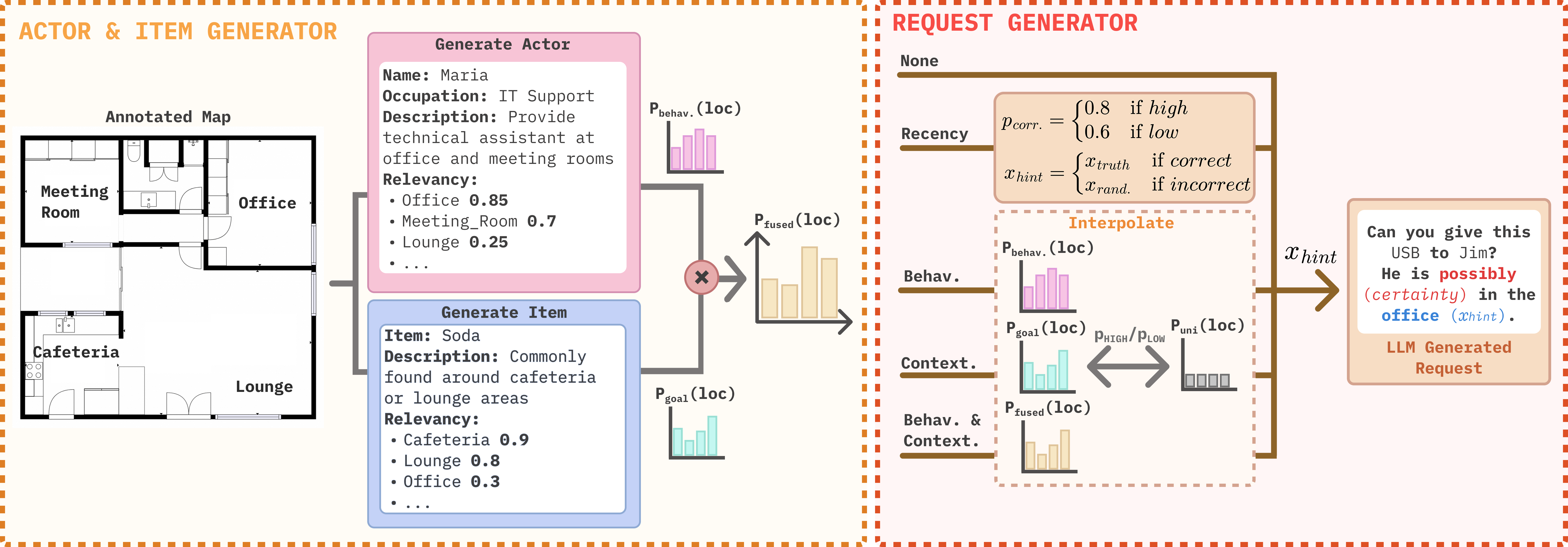}
    \caption{Synthetic Scenario (Actor, Item, and Request) Generator}
    \label{fig:scenario_gen}
\end{figure*}
\section{Scenario Generator}
\label{scenario_generator}

To evaluate the framework, a synthetic scenario generator is proposed to generate synthetic recipients, items to deliver, and requests defining each scenario. The request %information as mentioned in \ref{problem_formulation} is 
provides noisy natural language observation of the recipient's whereabouts as defined in Section \ref{problem_formulation}. %The request aimed to simulated plausible human interactions the agent will encounter with varying categories of trust and hints for evaluation and ablation studies. 
Requests simulate plausible human interactions with varying hint types and levels of certainty, enabling controlled evaluation and ablation studies. Since real-world requests are not available for this task, synthetic scenarios provide controlled ground truth for evaluating the methods prior to real-world deployment.% it's synthetically generated to quantify meaningful performance of the methods before real-world deployment.

\subsection{Environment Definition}
The environment is initialized from a floor plan of the building. This is annotated with an environment description (e.g. "This is an office environment") and discrete locations with labels and descriptors (e.g. "Lounge - this is a place for resting"). These annotations provide semantic information about each location that can be leveraged by an agent. This represents minimal information that can reasonably be made available in real-world deployment, requiring only a floor plan and one-time semantic annotation.%is also acts as minimal information provided to the agent as in the real-world floor plans are documented and can be available for use, alongside human annotation of the map only needing to be done once for real-world use.

\subsection{Actor and Item Generator}
\label{actor_generator}
The actors in the environment are the available recipients. Each actor is synthetically generated using an LLM that leverages the environment description to produce a name, role/occupation, and description of their behaviour, as shown in Figure \ref{fig:scenario_gen}. To define the recipient's behaviour prior, we generate a probability distribution of their behaviour, called the behavioural distribution \(p_{behav}(loc)\). This is done by using the LLM to also generate relevancy scores assigned to each location based on the description and occupation of the recipient similar to section \ref{belief_generator}. Items are generated using a similar process, producing the goal/intent distribution, \(p_{goal}(loc)\), which defines the probability of a general recipient needing to use that item for a given location. These two distributions model the individual ground truth based on recipient behaviour and recipient intent over the environment used to generate the request.

\subsection{Request Generator}
To generate the request itself the information required is the item, recipient, hint type, hinted location, and associated uncertainty. The item and recipient are randomly chosen. Their behavioural distributions, \(p_{behav}(loc)\) and \(p_{goal}(loc)\) are fused to obtain a fused distribution, \(p_{fused}(loc)\). This distribution represents the recipient's likely whereabouts given both their behavioural trends and the intent associated with the item. The recipient's ground-truth location is sampled from \(p_{fused}\), introducing controlled randomness while preserving plausible location preferences. %This is achieved by multiplying the distributions (amplifying agreeing ideas and diminishing disagreeing ones) and re-normalizing it to get a valid distribution. Sampling this distributions generates the ground truth location of the person (sampling adds biased randomness in accordance to plausible locations the recipient may be at using habits and intent to use the item).

The hint types (see Table \ref{tab:hint_types}) and associated uncertainty (HIGH or LOW) are randomly selected. Each hint type determines the information source used to generate the hinted location. None provides no additional location information. Recency is coupled directly to the sampled ground-truth location, selecting the true location with probability 0.8 for HIGH certainty and 0.4 for LOW certainty. Behavioural and Contextual hints are derived from \(p_{behav}\) and \(p_{goal}\), respectively, while Behav. \& Context. uses \(p_{fused}\). 

For these distribution-based hints, certainty is modelled by interpolating the corresponding distribution with a uniform distribution, using weights of 0.8 and 0.4 for HIGH and LOW certainty, respectively. The highest-probability location after interpolation is used as the hinted location. %The purpose of the hint is to act as an extra noisy indication of the recipients whereabouts providing a hinted location which different request sentence delivery to indicate the type of hint expressed. This is to evaluate agent performance on deciding what information to trust and to what degree. 
%For the None hint type this is a request without hinted information only including the base information (intent to use the item and recipient behaviour). As for recency since this is an observation over some time horizon modelled by coupling directly to the ground truth by giving an 80\% success rate of choosing the ground truth for lower uncertainty (e.g. "seen a few minutes ago") and a 40\% success rate for higher uncertainty (e.g. "seen a few hours ago"). Since recency is a factual observation with some uncertainty based on time the ground truth is directly coupled. For behavioural, contextual, and behav. \& context. hints theses these are inferred/estimated information from the user so they're sampled from their respective distributions with some applied interpolation to a uniform distribution similar to section \ref{belief_generator}. The behavioural and contextual use their respective marginal distributions \(p_{behav}(location)\) and \(p_{context}(location)\). Since behav. \& context. hint use both intent and routine of the recipient it's natural to use the combined distribution \((p_{fused}(location)\). Based on the uncertainty a interpolation percentage is chosen (percent\_high: 0.8 and percent\_low: 0.4) demonstrating how well calibrated the recipients latent understanding of the recipient is based on the hint type. The highest probable location in the generated distribution after interpolation is used as the hinted location to represent the user selecting the most probable location they think the recipient is at.
Once the item, recipient, hint type, uncertainty, and hinted location are determined, an LLM parses the information in a JSON schema to generate a natural language request as shown in Table \ref{tab:hint_types}.

\section{EXPERIMENTAL SETUP}
\subsection{Map Annotation And Synthetic Benchmark}

\begin{figure}[t]
    \centering
    \includegraphics[width=\linewidth]{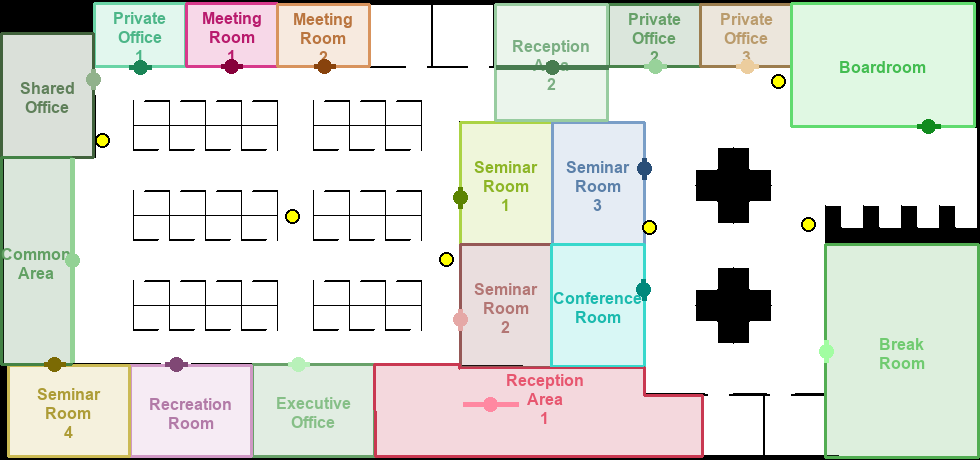}\\[0.2cm]
    \includegraphics[width=\linewidth]{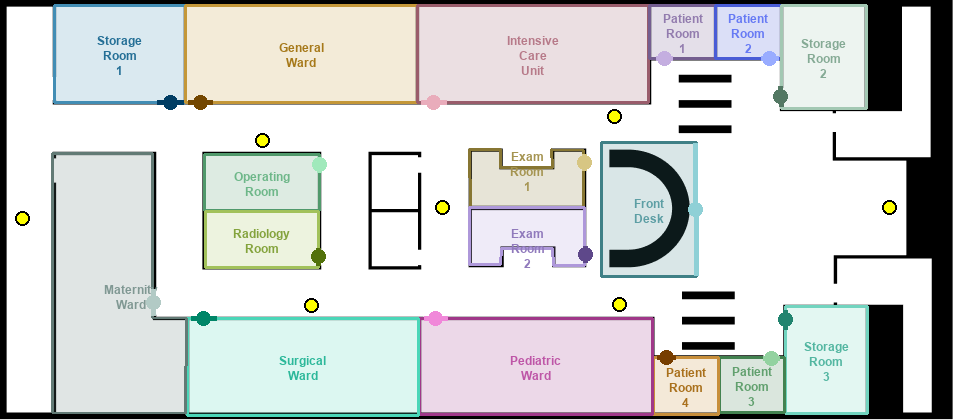}\\[0.2cm]
    \includegraphics[width=\linewidth]{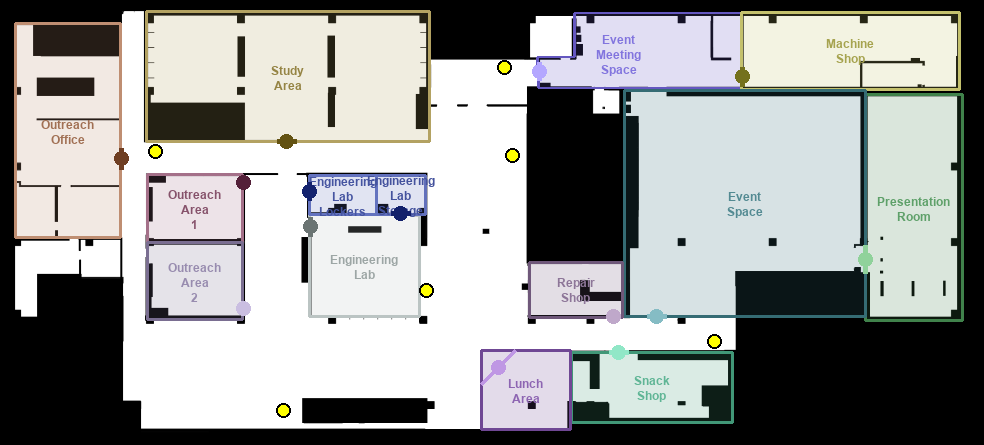}
    \caption{Annotated Office, Hospital, and University Map (yellow circle markers indicate agent start locations)}
    \label{fig:locs}
    %\caption{Annotated Office (yellow circles indicate agent start location)}
\end{figure}

The environments considered are: a generic office (18 loc.), a hospital (17 loc.), and an engineering university building (14 loc.). Each environment is annotated based on grounded and plausible locations labels and semantic descriptions shown in \ref{fig:locs}. The office and hospital are derived and labelled in accordance to \cite{fung2025mllm} that is spatially tied to each location. As for the engineering university building this is taken from an actual floor plan cleaned to be used as an occupancy map and labelled based on real-world locations in the building. These maps are chosen to reflect realistic environments the agent can be in for PersonNav tasks.

For each environment, 10 actors and 10 items are generated, with 500 scenarios evaluated for each environment for a total of 1,500 scenarios (containing 30 different actors and items). This provides coverage for ablation studies in terms of different hint types and also demonstrates diversity among semantic information (items, people, and environment). The LLM model for synthetic generation chosen is gpt-4o-mini (Temperature=1.7 and Top\_P=1.0) from the GPT suite of models \cite{sufi2024addressing}.

\subsection{Baselines}
These baseline methods range from classical graph-based planners to semantically informed planners to evaluate the contributions of semantic information, geometric distance, and route optimization. We consider zero-shot semantic methods rather than approaches requiring person-specific training data, consistent with our assumption of limited access to personal activity data. %to represent %fair evaluations against our method. Learned based methods are not introduced as one of the key considerations to the problem space is to prevent privacy leakage of human information for real world setting application therefore learned methods that would required data collection is not considered. Hence, zero-shot methods that are pre-trained are leveraged as baselines otherwise classical graph-based planners are used for this work. 
The baselines are as follows:

\begin{itemize}
    \item \textbf{Random}: randomly orders the candidate locations, providing a lower-bound baseline;% is the lower bound performance baseline
    \item \textbf{Greedy}: iteratively selects the closest unvisited location;%selects the next closest location until found
    \item \textbf{Held-Karp (TSP) \cite{held1962dynamic}}: computes the minimum-distance route through all candidate locations;% is a dynamic programming approach to solve the travelling salesman problem, optimizing for the total shortest route
    \item \textbf{Semantic Greedy}: orders locations according to their semantic probability;% selects the next highest probability
    \item \textbf{Semantic \& Distance}: leverages both distance and semantic belief in a heuristic \(p(loc)/d_{loc}(s)\) as the cost of probability per unit distance;
    \item \textbf{Semantic TRP \cite{afrati1986complexity}}: uses the semantic belief as location weights in the TRP objective to minimize cumulative cost (weighted by person finding probability).
\end{itemize}

\subsection{Metrics}

%These metrics define quantities such as the number of locations visited to the overall travel efficiency that addresses objective 1.
Performance is evaluated using metrics that capture search accuracy, travel efficiency, planning cost, and belief quality:

\begin{itemize}
    \item \textbf{Overlap Coefficient} is used to evaluate the benchmark and is used as a distance measure between the behavioural and contextual distributions. \(overlap(A, B) = \frac{|A \cap B|}{min(|A|,|B|)}\) \cite{inman1989overlapping}
    \item \textbf{Top-K Accuracy} is the probability of finding the person within K-visited locations
    \item \textbf{Visited Locations} is the average number of locations travelled for a given environment
    \item \textbf{Travel Efficiency} is the ratio between the shortest path to the actual path (closer to 1 is optimal). \(T_{eff} = d_{shortest}/d_{actual}\)
    \item \textbf{Planning Time} is total time (seconds) it takes from receiving a request to generating a valid route.
    \item \textbf{Kullback-Leibler Divergence \cite{kullback1951information}} defines the amount of information lost in distribution Q with respect to ground truth distribution P. This represents how well the generated belief maintains information from the true fused distribution. \(D_{KL}(P||Q) = \sum P(X) log(\frac{P(X)}{Q(X)})\)
\end{itemize}

\begin{table*}[t]
\centering
\caption{Simulation Summary Performance Accross All Environments}
\label{tab:results}
\resizebox{\textwidth}{!}{%
\begin{tabular}{l cccc cccc cccc ccc}
\toprule
& \multicolumn{4}{c}{\textbf{Office}} & \multicolumn{4}{c}{\textbf{Hospital}} & \multicolumn{4}{c}{\textbf{Engineering University}} & \multicolumn{3}{c}{\textbf{Average}} \\
\cmidrule(lr){2-5} \cmidrule(lr){6-9} \cmidrule(lr){10-13} \cmidrule(lr){14-16}
Method & Travel Eff. & Top-3 Acc. & Visited & Plan Time & Travel Eff. & Top-3 Acc. & Visited & Plan Time & Travel Eff. & Top-3 Acc. & Visited & Plan Time & Travel Eff. & Top-3 Loc. & Plan Time. \\
\midrule
Graph-Based Planners\\
\midrule
Random          & 0.22 & 0.17 & 9.42  & -& 0.22& 0.19& 7.75& -& 0.21& 0.20& 8.92& -& 0.22& 0.19&- \\
Greedy          & 0.36 & 0.13 & 10.53 & -& 0.48& 0.20& 7.22& -& 0.43& 0.18& 10.10& -& 0.42& 0.17&- \\
Held-Karp TSP   & 0.41 & 0.25 & 9.80  & \cellcolor{gold!40}\textbf{0.11}& 0.44& 0.16& 7.56& \cellcolor{gold!40}\textbf{0.01}& 0.44& 0.20& 9.81& \cellcolor{gold!40}\textbf{0.05}&0.43 & 0.20&\cellcolor{gold!40}\textbf{0.06} \\
\midrule
Semantic Planners\\
\midrule
LLM             & 0.52 & \cellcolor{gold!40}\textbf{0.53} & \cellcolor{gold!40}\textbf{4.80}  & 3.09& 0.59& \cellcolor{gold!40}\textbf{0.7}& \cellcolor{gold!40}\textbf{3.12}& \cellcolor{silver!40}1.55& 0.46& \cellcolor{gold!40}\textbf{0.52}& \cellcolor{gold!40}\textbf{4.75}& \cellcolor{silver!40}1.84&0.52 &\cellcolor{gold!40}\textbf{0.58} &\cellcolor{bronze!40}2.16 \\
LLM \& Dist.     & \cellcolor{bronze!40}0.57 & \cellcolor{silver!40}0.44 & \cellcolor{silver!40}5.74  & \cellcolor{silver!40}1.90& \cellcolor{silver!40}0.65& \cellcolor{silver!40}0.58& \cellcolor{silver!40}3.87& 1.90& \cellcolor{bronze!40}0.51& \cellcolor{silver!40}0.41& 5.82& 2.33& \cellcolor{bronze!40}0.58& \cellcolor{silver!40}0.48&6.13 \\
LLM TRP         & \cellcolor{silver!40}0.60 & 0.25 & 7.44  & \cellcolor{bronze!40}2.10& \cellcolor{bronze!40}0.63& 0.39& 5.11& 1.78& \cellcolor{silver!40}0.56& 0.31& 6.47& \cellcolor{bronze!40}2.07& \cellcolor{silver!40}0.60&0.32 &\cellcolor{silver!40}1.93 \\
\midrule
\textbf{CSIR (Ours)} & \cellcolor{gold!40}\textbf{0.63} & \cellcolor{bronze!40}0.30 & \cellcolor{bronze!40}6.68 & 3.92 & \cellcolor{gold!40}\textbf{0.67}& \cellcolor{bronze!40}0.56& \cellcolor{bronze!40}3.96& 3.93& \cellcolor{gold!40}\textbf{0.57}& \cellcolor{bronze!40}0.34& \cellcolor{bronze!40}6.38 & 4.07& \cellcolor{gold!40}\textbf{0.62}&\cellcolor{bronze!40}0.4 &3.98 \\
\bottomrule
\end{tabular}%
}
\end{table*}
\section{Synthetic Benchmark Characterization}
\label{sec:sythn_eval}
\begin{figure}[h]
    \centering
    \includegraphics[width=1\linewidth]{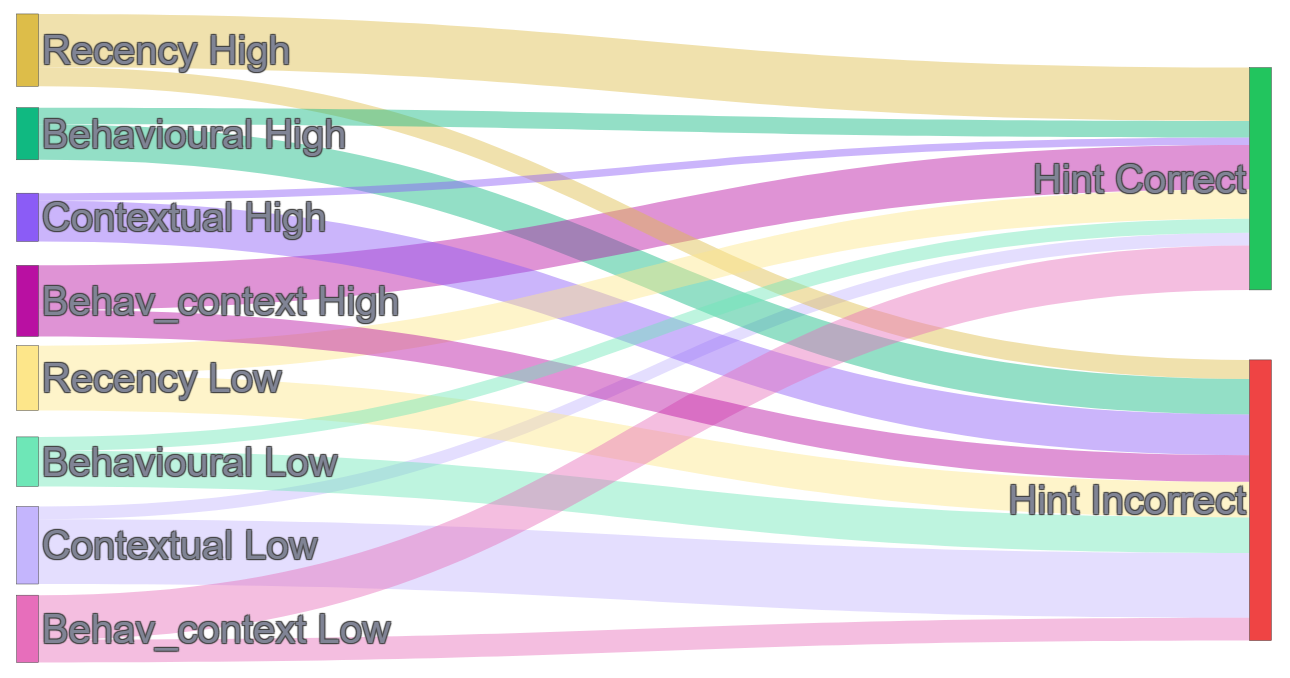}
    \caption{Sankey Diagram to Show Hint and Certainty Distribution to Hint Correct/Incorrect (Office Environment)}
    \label{fig:sankey}
\end{figure}

%After generating the benchmark and scenarios the requests are evaluated to see for any consistencies within the benchmark. 
We first characterize the generated scenarios to evaluate whether the hint types and certainty levels reflect the intended properties of the benchmark. Fig. \ref{fig:sankey} shows the relationship between hint type, certainty, and whether the hinted location matches the recipient's ground truth location. HIGH-certainty hints are expected to provide more reliable indications of the recipient's location than LOW-certainty hints. %this displays all the hint and certainty permutations showing the distribution of how much it's a correct or incorrect hint. The initial hypothesis is that for higher certainty hints they should be biased towards getting the hint correct and vice versa for lower certainty hints. 
This is most apparent for Recency and Behav. \& Context. hint types. 
Recency is directly coupled to the samples ground-truth location, while Behav. \& Context. is derived from the fused behavioural and goal distribution used to generate the recipient's location. Consequently, these hint types are more closely related to the benchmark ground truth. One discrepancy is that Behav. \& Context. hint type shows a bias towards the correct hint under low certainty. This is due to the fused distributions having a narrow peak meaning the marginal distributions are not high in agreement further explained below.

%This falls back to how the sampling for the hinted location is achieved and aligns with what is designed in Section \ref{}. This is because recency and behav. \& context are grounded with respect to the true location and true fused distribution respectively. This means the users observations are more grounded reflecting that last seen locations and predictions from both routine and intent information are grounded information markers. 

Contextual and Behavioural hint types, in contrast, are derived from the marginal distributions of the fused distribution. Their agreement with the ground-truth location therefore depends on how strongly each individual distribution aligns with the fused distribution \(p_{\mathrm{fused}}\). This reflects the benchmark design, where individual sources provide partial information about the recipient's location.
%this purely depends on how well the distributions align or overlap, since if the distributions are quite different sampling from it will not contribute much towards the correct location hint. This can be understood that the benchmark is designed so that isolated pieces of information are not strong enough markers of the fused distribution hence the grounding to the true location is weak. This is more so a design choice as isolated or sparse information as your basis can yield weaker predictions.

The overlap coefficient is also computed for the Behavioural and Contextual hints showing how well they overlap. This averages out to 0.32 (Office: 0.27, Hospital: 0.43, University: 0.25.). The relatively low overlap indicates that the two information sources frequently favour different locations, allowing the benchmark to evaluate search under complementary and potentially conflicting information.%distributions do not overlap enough to make meaningful gains as a marker of the fused distribution but still can be reflective and meaningful to benchmark how weaker sources of information impact performance.

\section{CSIR Framework Evaluation}
\subsection{Uncertainty Predictor Results}
% The certainty predictor uses a semantic model (transformer encoder-type models) that generates a vector (1024 x 1) which is then trained with a linear or non-linear head to output a single value \(\alpha\). The models evaluated are as shown in table \ref{tab:certainty_results} as open-sourced models available for fine-tuning. As mentioned in section \ref{belief_generator} the dataset to fine-tune on is generated synthetically using gpt-4o-mini. This generates statements with varying level of speaker certainty (also known as hedging language or epistemtic modality) with buckets of labels that also have a quantity of uncertainty [0, 1]. The dataset is generated with 2000 labels of data and is human reviewed to check for consistency across generated data. The models are trained over this dataset and the validation accuracy is reported in table \ref{tab:certainty_results} using a 80:20 train validation split. This is seperate from the synthetic scenario benchmark to allow the model to generalize uncertainty phrases regardless if it takes on the form of a request. To further see how well the models perform an unseen dataset it's then evaluated over the synthetic scenario benchmark on the requests itself.

The uncertainty predictor uses a BERT-like model \cite{gardazi2025bert} which produces an embedding vector (1024 x 1) and a probe (or multi-layer head) to output a single continuous value, \(\alpha\). Different open-source models are evaluated as shown in Table \ref{tab:certainty_results}. The dataset mentioned in \ref{belief_generator} is generated using gpt-4o-mini synthetically generating 2000 statement and label pairs  that are reviewed for consistency. The models are fined-tuned using a 80:20 training and validation split. Table \ref{tab:certainty_results} shows the classification accuracy on the validation set and also the generated requests (from the 1,500 scenarios), evaluating how well the models generalize to sentences with varying levels of uncertainty.

% Fine Tuned Office 0.654, Uni 0.712, Hospital 0.648
% Linear Office 0.811, Uni 0.8430, Hospital 0.796
% Non-Linear Office 0.914, Uni 0.954, Hospital 0.940
\begin{table}[h]
    \centering
    \begin{tabular}{llcc}\toprule
         \textbf{Model} &\textbf{Train Method}&  \textbf{Val. Acc.}&  \textbf{Scenario Acc.}\\\midrule
         DeBERTa-v3-base  &Fine-Tuning&  0.86&  0.67\\
         Qwen3-0.6B-\\Embedding &Linear Head&  0.80&  0.82\\
        Qwen3-0.6B-\\Embedding &Non-Linear Head& \cellcolor{blue!40}\textbf{0.87}& \cellcolor{blue!40}\textbf{0.94}\\ \bottomrule
    \end{tabular}
    \caption{Certainty classification accuracy on validation and scenario datasets}
    \label{tab:certainty_results}
\end{table}

% \begin{figure}[h]
%     \centering
%     \includegraphics[width=1\linewidth]{figures/umap.png}
%     \caption{Latent Space Representation of the Sequential Embedding Model (Learned Categorization of Uncertainty)}
%     \label{fig:umap}
% \end{figure}

The results show that the Qwen-3-0.6B-Embedding models \cite{Qwen} capture linguistic uncertainty and hedging language more accurately compared to the BERT model. The non-linear head achieves the highest accuracy, with 0.87 validation accuracy and 0.94 scenario accuracy. %used on the embedding model is able to better separate the data showing significant performance improvement by 12\%. 
%The model %is also able to generalize well on the synthetic scenario dataset meaning it's able to properly extrapolate what language uncertainty means in its latent space, this separation can be seen in 
% also shows increased separation between certainty classes after training, as illustrated in Fig. \ref{fig:umap}. 
This performance increase from the training dataset to scenario dataset is likely due to the classification bucket \(\alpha\) being reduced from five to two (HIGH and LOW) in the scenario dataset.

\subsection{Simulation Results}
\label{sim_result}
The results show the full evaluation on the synthetic benchmark in Table \ref{tab:results}. The random case provides the lower bound reference for performance in the scenarios. The top-3 accuracy and visited location results show that semantically informed planners perform better than classical/graph-based planners in terms of finding the recipient. This shows that informing search adds meaningful information in reducing the number of candidate places to explore. As for travel efficiency the LLM greedy and LLM \& Dist. show improvements over the graph-based planners demonstrating semantic information can be leveraged to inform search even under varying informational uncertainty. Incorporating distance and TRP-based route optimization further improves average travel efficiency, with LLM \& Dist., LLM TRP, and CSIR (Ours) achieving 0.58, 0.60, and 0.62, respectively. %Since these semantic methods rely on heuristic methods the TRP solutions LLM TRP and Ours does progressively better (2\%-8\% improvement in travel efficiency). 
The similarity between LLM TRP and Our method shows that the belief generator produce similar search performance. Although their performance is comparable, the LLM TRP method abstracts how it numerically generates its distribution making it difficult to reason about how each different information source contributes to the generated belief. Our method instead explicitly represents the individual information distributions, allowing their uncertainty to be inspected and modulated independently (e.g. if the assumption that item or occupation information is factual no longer holds).

The results across different environments for a given method shows relatively similar performance across the board showing that such semantic methods are robust against different environmental changes.

% ^ Trying to justify why our framework is still meaningful even if comparable to a prompted engineered LLM solution (by having actual distributions we can calibrate these unlike emperical prompt tuning) (Does slightly better in terms of performance but not significantly)

% it goes to show by discretely generating the distributions involved in generating the initial belief is beneficial for future improvements and even using Bayesian-style filtering for encountered uncertain observations (e.g. environment changes like doors open and closed can be uncertain indicators of human activity). This shows that our method is still valid and the correct objective is chosen for the problem optimizing for the belief weighted cumulative distance cost for each location.

\begin{figure}[h]
    \centering
    \includegraphics[width=1\linewidth]{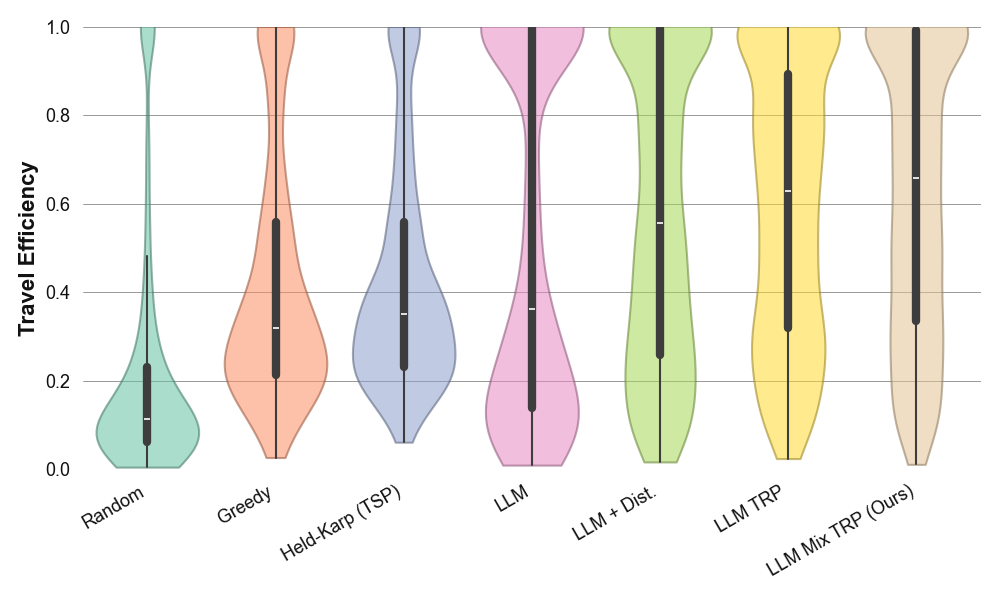}
    \caption{Violin Plot of Travel Efficiency Results Across All Methods (All Environments)}
    \label{fig:violin}
\end{figure}

Fig. \ref{fig:violin} shows the results for the travel efficiency display the full distribution as a violin plot. The resulting distributions shown are bimodal, with the concentration of mass near the extremes (high and low travel efficiencies). This indicates that some methods may find the recipient either near the beginning or at the end of the search route. Comparing LLM only (belief only method) to the other belief and distance fused methods (LLM \& Dist., LLM TRP, and Ours) shows that this bimodality is reduced and concentrates towards the higher extreme. This suggests that semantic information can effectively inform search but may produce inefficient routes when the inferred belief is inaccurate. 
By using distance in the planned route, the search is spatially grounded, reducing sporadic exploration based on a noisy initial belief.%spatially grounds the search space to prevent sporadic exploration based on a noisy initial belief, hence the failure mode seen by a semantic only method is reduced significantly (shown in the plot pushing the mass towards the higher end of the plot).

To address the comparison of heuristic and exact solutions, the planning time in Table \ref{tab:results} shows that computation remains practical for environments with up to 18 locations. However, for environments with more than 27 locations, planning takes over one minute and requires several GB of RAM, making the exact solution less suitable for time-constrained applications or robots with minimal computational power. Hence, other heuristic methods such as LLM \& Dist. represent a meaningful trade-off.

% This can be seen for the fact that if the method is unable to find the recipient it will fall back to a distance based traversal superior to a random or noisy search (sporadically searching locations far apart).

\subsection{Belief Generator Ablation}

This ablation study looks at the performance of different belief generators including the bi-encoder (BERT-based semantic model), cross-encoder (BERT-based semantic model), LLM, and CSIR (Belief, different from Table \ref{tab:results} since no route planner involved). Both the bi and cross encoder use cosine similarity between the request vector and the information vector (containing environment labels and recipient occupation) to generate a score for each location and is normalized after. 

\begin{table}[h]
    \centering
    \begin{tabular}{lcccc}\toprule
         Belief Gen.&  Travel Eff. &Top-3 Acc.&  KL Div.& Gen. Time\\\midrule
         Bi-Encoder&  0.38&0.37&  1.55& 0.03\\
         Cross-Encoder&  0.48&0.47&  1.86& \cellcolor{blue!40}\textbf{0.01}\\
         LLM&  \cellcolor{blue!40}\textbf{0.52}&\cellcolor{blue!40}\textbf{0.58}&  \cellcolor{blue!40}\textbf{1.36}& 2.16\\
 CSIR (Belief)& 0.51&0.56& 1.58&3.70*\\ \bottomrule
    \end{tabular}
    \caption{Belief Generator Ablation Results (*Note: Gen. time is computed using the hospital and university results since parallelization was not introduced for the office yet)}
    \label{tab:belief_ab}
\end{table}

The results show that the LLM achieves the highest performance, with CSIR (Belief) performing similarly. This demonstrates that LLM methods outperform BERT-based models but at higher the cost of inference time (2 to 4 seconds difference). The performance similarity between LLM and CSIR (Belief) supports the observation in Section \ref{sim_result} that explicitly decomposing the belief maintains comparable performance while enabling individual information sources to be calibrated or updated independently. % where even similar our method is able to properly extract out the information numerically into meaningful distributions for calibration or filtering methods on the sub information level.

% This shows that LLMs are able to better reason between why the information and request correlate. The cross and bi encoder do perform much faster as seen in \ref{tab:belief_ab} but may trade-off in terms of performance since cosine similarity provides semantic similarity of sentences but not deeper coherent reasoning provided in the larger models (stronger prior information).

% Hos, Uni, Office
%Bi-Encoder	0.1959+0.5612+0.3844	0.152+0.592+0.36	1.3553+1.4092+1.8855	0.0333+0.02857+0.033
%Cross-Encoder	0.4068+0.5480+0.4795	0.386+0.536+0.484	1.6349+1.7686+2.1653	0.0126+0.0101+0.01158

\subsection{Hint Type Ablation}

The hint type ablation evaluates how each information type affects the performance of our method, as shown in Tab. \ref{tab:hint_ab}.

\begin{table}[h]
    \centering
    \begin{tabular}{lcc}\toprule
         Hint Type&  Travel Eff.&  KL Div.\\\midrule
         None&  0.46&  1.63\\
         Recency&  0.80&  1.49\\
         Behavioural&  0.67&  1.80\\
 Contextual& 0.69& 1.58\\ 
 Behav. \& Context.& 0.77& 1.42\\
 \bottomrule
    \end{tabular}
    \caption{Hint Type Ablation Results}
    \label{tab:hint_ab}
\end{table}

The None hint produces the lowest travel efficiency, which aligns with expectations as the framework has no additional information outside of the item and recipient occupation. Its KL divergence of 1.63 is also relatively high, indicating greater divergence from the ground-truth distribution. 
Recency and Behav. \& Context. achieve the highest travel efficiencies (also shown with the lowest KL divergences), at 0.80 and 0.77, corresponding to increases of 0.34 and 0.31 over the None condition. This is consistent with the benchmark design, where these hints are coupled with the true location and fused distribution. Contextual and Behavioural hints also improve travel efficiency to 0.69 and 0.67, respectively.  Despite their weaker correspondence with the ground-truth location discussed in Section \ref{sec:sythn_eval}, both outperform the None condition. This may partly result from the route planner incorporating distance, thereby reducing sporadic or random exploration.%These results are surprising since as mentioned in section  the hint types are poor predictions of the grounded truth mostly resulting in wrong hints. Even if the hinted locations are mostly wrong it can still perform better than the None hint type capable of extracting relevant information for person finding. This can also be attributed to the route planner ensuring that even wrong initial beliefs are compensated using distance for search reducing sporadic or random exploration.

% This can also be attributed to the optimal approach for route planning as a plausible source of performing reasonably under mostly incorrect information. These results show the framework is able to meaningfully extract information and ground itself during route planning to perform under reasonable conditions regardless if whether information is correct or incorrectly provided.

\subsection{Hardware Results}
\begin{figure}[!h]
    \centering
    \includegraphics[width=1.0\linewidth]{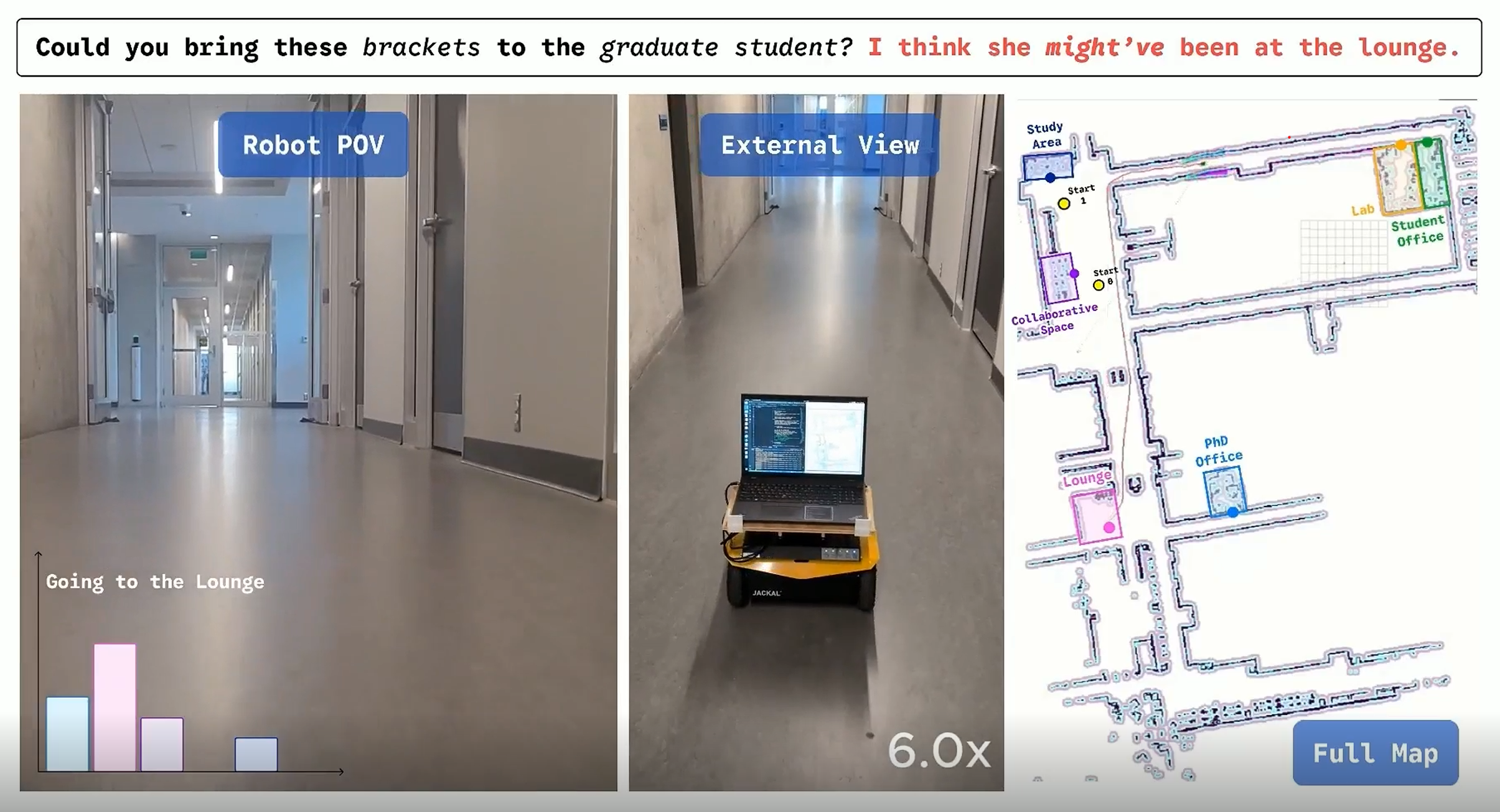}
    \includegraphics[width=1.0\linewidth]{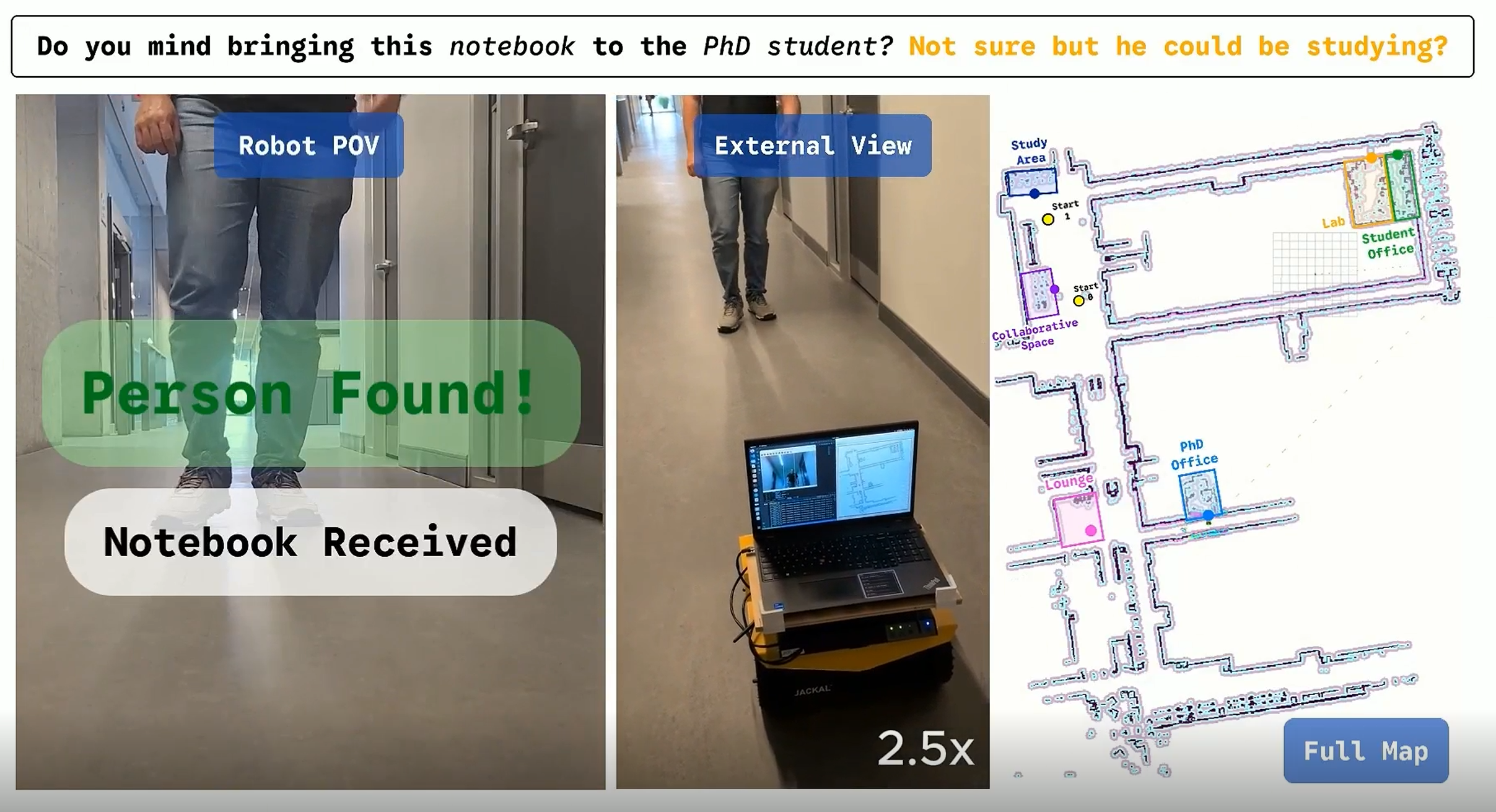}
    \caption{Hardware Experiment Recency and Behavioural Hint Requests}
    \label{fig:placeholder}
\end{figure}
For real-world embodiment, the framework was implemented on the Clearpath Jackal \cite{ClearpathRobotics_2026} to perform item delivery tasks in a real university building environment. An earlier iteration of the framework was used, as hardware testing preceded the subsequent framework improvements; these experiments therefore demonstrate real-world feasibility rather than evaluate the final framework. To generate the annotated map a LiDAR scan of the floor is used and annotated based on real rooms and spaces. The recipients were graduate students, with requests designed to reflect their plausible behaviours. Since the framework is a planning method for actual exploration Nav2 \cite{NAV2} is used with key way points at doors defined as a location coordinate. A RealSense D435 camera is used for RGB images to run YOLO \cite{jiang2022review} for detecting a person and DeepFace models \cite{taigman2014deepface} for face detection to determine if the correct recipient was found. 
The experiments showed that the robot was able to complete the task within approximately 5 minutes, using both semantic and distance information to guide its search. The experiments also highlighted some real-world constraints such as accessibility (elevators, doors, private areas) and local search costs within larger rooms, which motivate extensions of the framework in future work. \textbf{\textcolor{red}{\href{https://anonymous.4open.science/r/personnav-site-4ED2/index.html}{https://anonymous.4open.science/r/personnav-site-4ED2/index.html}.}}

\section{CONCLUSIONS}
This work aimed to address the Person Goal Navigation problem under real-world constraints on information accessibility and uncertainty. The proposed framework enables efficient PersonNav in structured indoor environments by combining distance and semantic scores weighted by the level of trust of information provided by the user. For evaluation, a benchmark was generated to synthetically create actors, items, and requests with controlled ground truth across multiple environments. The results show that semantically informed search improves over classical graph-based baselines based purely on distance. However, semantics alone can lead to inefficient search when not grounded by distance. CSIR achieves the highest average travel efficiency among the evaluated methods (0.62), while performing comparably to LLM TRP (0.60). Our method additionally provides greater transparency over the underlying distributions that contribute towards the final generated belief allowing individual information sources to be independently calibrated or updated. Future work will focus on refining the generated belief using Bayesian filtering and observations acquired during search, and consider additional travel costs such as room-based exploration. These extensions will further support informed PersonNav frameworks under realistic constraints for mobile service robots.%This work aimed to develop more intelligent mobile servicing agents to alleviate tasks in item-delivery scenarios allowing for informed search in environments reflect realistic constraints and allow for robots to leverage natural language observations to inform decisions in the human world.

% But, with out method providing more clarity of the distributions used to generate the belief Bayesian-filtering methods can be easily appended for future work. Future work will focus on improving the framework to contain better representations of belief, use information found during search to perform Bayesian filtering, and consider additional costs such as local search and exploration on the room-level. This work aimed to develop more intelligent mobile servicing agents to alleviate tasks in item-delivery allowing for informed search in environments reflect realistic constraints and allow for robots to leverage human-like observations to inform decisions in the human world.

% \section{ACKNOWLEDGEMENT}
% We acknowledge the support of the Natural Sciences and Engineering Research Council of Canada (NSERC), [funding reference number CREATE 584985-2024].

\bibliographystyle{IEEEtran}
\typeout{}
\bibliography{reference.bib}

\end{document}